\documentclass[runningheads]{llncs}

\usepackage{eccv}

\usepackage{eccvabbrv}

\usepackage{graphicx}
\usepackage{booktabs}
\usepackage{amsmath}
\usepackage{amssymb}
\usepackage{multirow}
\usepackage{xcolor}
\usepackage{colortbl}
\usepackage{makecell}
\usepackage{subcaption}
\definecolor{seagreen}{RGB}{46, 139, 87}

\usepackage[accsupp]{axessibility}  % Improves PDF readability for those with disabilities.

\usepackage{hyperref}

\usepackage{orcidlink}

\newcommand{\ours}{\textsc{CoLT}}
\newcommand{\Lfwd}{\mathcal{L}_{\text{fwd}}}
\newcommand{\Lbwd}{\mathcal{L}_{\text{bwd}}}
\newcommand{\Lint}{\mathcal{L}_{\text{int}}}
\newcommand{\Ltask}{\mathcal{L}_{\text{task}}}
\definecolor{bestcolor}{rgb}{0.85,0.92,0.85}
\definecolor{Gray}{gray}{0.88}
\begin{document}

% ---------------------------------------------------------------
\title{CoLT: Teaching Multi-Modal Models to Think\\with Chain of Latent Thoughts} 

\titlerunning{CoLT: Chain of Latent Thoughts}

\author{Lianyu Hu\inst{1}\orcidlink{0000-0003-2470-8110} \and
Shengqian Qin\inst{2}\and
Zeqin Liao\inst{1} \and
Qing Guo\inst{3,4}\thanks{Corresponding author} \and
Liang Wan\inst{5} \and
Wei Feng\inst{5} \and
Yang Liu\inst{1}}

%\author{Lianyu Hu\inst{1}\orcidlink{0000-0003-2470-8110} \and
% Shengqian Qin\inst{2}\orcidlink{1111-2222-3333-4444} \and
% Zeqin Liao\inst{1}\orcidlink{2222--3333-4444-5555} \and
% Qing Guo\inst{3}\orcidlink{0000-0003-0974-9299} \and
% Liang Wan\inst{4}\orcidlink{2222--3333-4444-5555} \and
% Wei Feng\inst{4}\orcidlink{0000-0003-3809-1086} \and
% Yang Liu\inst{1}\orcidlink{2222--3333-4444-5555}}
\authorrunning{Lianyu Hu et al.}
\institute{Nanyang Technological University, Singapore \and Shanghai Jiao Tong University, China \and NKIARI, Shenzhen Futian, China \and  AAIS \& VCIP, Nankai University, China \and Tianjin University, China}

\maketitle

\begin{abstract}
Chain-of-thought (CoT) reasoning has enabled multi-modal large language models (MLLMs) to tackle complex visual reasoning tasks by generating explicit intermediate reasoning steps in natural language. However, this text-based reasoning paradigm is inherently slow at inference time with even thousands of tokens and fundamentally constrained by the expressiveness of natural language. In this paper, we propose \ours{} (\textbf{C}hain \textbf{o}f \textbf{L}atent \textbf{T}houghts), a novel framework that teaches multi-modal models to reason through a chain of latent thought representations instead of verbose text tokens, which can perform thinking with as few as 3 steps. Naively forcing the model to think with latent states easily produces meaningless semantics and makes training unstable. To effectively regulate the latent reasoning process, we introduce a lightweight external decoder that provides step-level supervision for each latent reasoning step in two complementary directions: a \emph{forward} mode that decodes latent thoughts into the textual reasoning of the next step, and a \emph{backward} mode that aligns decoder hidden states with the model's latent thoughts given preceding textual context. We further incorporate \emph{internal supervision} that encourages coherent step-by-step latent transitions. The decoder and internal supervision are removed during inference to maintain high efficiency of latent reasoning.  Extensive experiments on eight benchmarks demonstrate that \ours{} not only outperforms existing latent reasoning methods such as CODI and SIM-CoT, but also surpasses latent visual reasoning approaches that rely on auxiliary images with costly annotation requirements. Compared to text CoT methods, \ours{} can notably reduce the inference time by 10.1$\times$ and text decoding time by 22.6$\times$. Code is released at \url{https://github.com/hulianyuyy/CoLT}.

\keywords{Latent reasoning \and Multi-modal models \and Chain-of-thought}
\end{abstract}

\section{Introduction}
\label{sec:intro}
Multi-modal large language models (MLLMs) have demonstrated remarkable capabilities in visual understanding and reasoning~\cite{lyu2026photocraft,zhu2025internvl3,bai2025qwen3,huang2026step3,wang2025diffusion,li2024clip,shi2026intelligent,hu2026illava,hu2026tvi}. By integrating vision encoders with powerful language model backbones, these models can process both visual and textual information to solve complex tasks ranging from visual question answering to mathematical problem solving~\cite{lu2024mathvista,yue2024mmmu}. A key ingredient underlying their success is chain-of-thought (CoT) reasoning~\cite{wei2022chain}, which instructs models to decompose problems into intermediate reasoning steps expressed in natural language before arriving at a final answer. Recent reasoning models such as DeepSeek-R1~\cite{guo2025deepseek} and QwQ~\cite{qwq2024} have further amplified this paradigm through long-form CoT, establishing it as the dominant reasoning strategy.

Despite its effectiveness, text-based CoT suffers from several fundamental limitations. First, generating lengthy reasoning chains in natural language significantly increases inference latency, as each reasoning token must be produced autoregressively. This becomes more severe as the thinking process typically consumes hundreds or even thousands of tokens~\cite{jaech2024openai,guo2025deepseek,huang2026step3}, constituting the heaviest computing burden during inference. Second, the decoded textual expressions are inherently deterministic: once a reasoning step is committed to a specific natural language form, it collapses the rich distribution of potential solving trajectories into a single surface realization, eliminating alternative reasoning pathways. Third, errors made in early reasoning steps can propagate and compound through subsequent steps, a phenomenon that is particularly problematic in long reasoning chains~\cite{wang2022self,li2025lvr}.

Recent work has begun to explore \emph{latent reasoning} as an alternative, where the intermediate thinking process occurs in a continuous representation space rather than through explicit text generation. CoCoNut~\cite{hao2024coconut} pioneered this direction by training language models to reason in a continuous latent space. CODI~\cite{shen2025codi} further explored compressing chain-of-thought into continuous representations via self-distillation. SemCoT~\cite{he2025semcot} tried to align latent thoughts with textual CoT via a trained sentence transformer. However, these approaches are primarily designed for text-only language models and do not address the unique challenges of multi-modal reasoning. More recently, latent visual reasoning~\cite{sun2025lacot,li2025lvr} leverages auxiliary images to generate latent thought tokens, but introduces significant overhead by requiring additional image annotations and specialized encoders, resulting in high labelling costs and limited scalability.

In this paper, we propose \ours{} (\textbf{C}hain \textbf{o}f \textbf{L}atent \textbf{T}houghts), a novel framework that enables multi-modal models to perform effective latent reasoning with principled supervision. A critical challenge is that naively forcing the model to perform latent thinking, \ie, simply replacing text tokens with continuous vectors during reasoning, easily produces semantically meaningless latent representations and causes severe training instability, as the unconstrained continuous space provides no inductive bias for structured reasoning. To overcome this, we introduce fine-grained step-level supervision through a lightweight external decoder operating in two complementary modes: 

\begin{itemize}
    \item \textbf{Forward decoding supervision}: The decoder takes the preceding latent thoughts as input and is trained to decode the textual reasoning content of the next step, ensuring that latent thoughts encode meaningful and progressive reasoning information.
    \item \textbf{Backward decoding supervision}: The decoder receives the preceding textual thoughts as input, and the generated hidden states are enforced to approach the latent thought representations, thereby anchoring the latent space to the semantic structure of explicit reasoning.
\end{itemize}

In addition to these external supervision signals, we introduce an \textbf{internal supervision} mechanism that directly encourages coherent transitions between consecutive latent steps by training the model to predict the next step's representation from the current one. The combination of forward, backward, and internal supervision provides a comprehensive training signal that effectively regularizes the latent reasoning process without requiring any auxiliary visual annotations.

Extensive experiments on eight multimodal benchmarks demonstrate that \ours{} consistently outperforms existing latent reasoning methods, and surpasses latent visual reasoning approaches that rely on costly image-based supervision. Compared to text CoT methods, CoLT can notably reduce the inference time by 10.1$\times$ and text decoding time by 22.6$\times$, achieving superior efficiency. Plentiful visualizations verify that \ours{} can successfully encode meaningful thinking patterns within the latent thoughts.

\section{Related Work}
\label{sec:related}

\subsection{Multi-Modal Large Language Models}

The rapid advancement of large language models has catalyzed the development of multi-modal large language models (MLLMs) that integrate visual perception with language understanding~\cite{liu2024llava,zhu2025internvl3,zhu2026medeyes,xiao2026not,zhu2025pathology,zhu2026medsynapsevbridgingvisualperception,bai2026tgo,bai2026prism,chen2024internvl25,bai2025qwen3,huang2026step3,fu2026neurosymactive,fu2026s}. LLaVA~\cite{liu2024llava} pioneered the visual instruction tuning paradigm, and its successors have significantly advanced the field: LLaVA-NeXT~\cite{liu2024llavanext} improved reasoning and OCR capabilities, while LLaVA-OneVision~\cite{li2024llavaonevision} unified image, video, and multi-image understanding into a single framework. The InternVL series has pushed open-source performance, with InternVL 2.5~\cite{chen2024internvl25} achieving results competitive with commercial systems, and InternVL3~\cite{zhu2025internvl3} further advancing capabilities via mixed-modality pre-training. In the Qwen-VL family, Qwen2-VL~\cite{wang2024qwen2vl} introduced native dynamic resolution support, and Qwen3-VL~\cite{bai2025qwen3} strengthened document parsing and long-video comprehension. Step3-VL~\cite{huang2026step3} conducted a vision-centric exploration of MLLM design choices, providing insights into the role of visual representations. While these models have made substantial progress, their reasoning capabilities still heavily rely on text-based intermediate representations, which our work aims to address through latent reasoning.

\subsection{Chain-of-Thought Reasoning}

Chain-of-thought (CoT) prompting~\cite{wei2022chain} has emerged as a powerful technique that enables language models to solve complex problems by generating intermediate reasoning steps. Li \etal~\cite{zhang2024chain} showed that CoT empowers transformers to solve inherently serial problems that are otherwise intractable for constant-depth models. Recent work has dramatically scaled this paradigm: OpenAI o1~\cite{jaech2024openai} demonstrated that training with long chains of thought yields substantial gains on scientific and mathematical benchmarks, while DeepSeek-R1~\cite{guo2025deepseek} showed that reinforcement learning can incentivize emergent reasoning capabilities. More recent works like OpenAI GPT-5~\cite{singh2026openai}, GLM-5~\cite{zeng2026glm} and Qwen2.5-xCoder~\cite{yang2025qwen2} further verify that long-form CoT is the key protocol to build high-intelligence agents.

In the multi-modal domain, CoT reasoning has been extensively explored across diverse modalities. Image analysis~\cite{shao2024visual,xu2025llava}, video understanding~\cite{zhang2025video} and embodied intelligence~\cite{zhao2025cot,ye2025vla} have fully witnessed the power of CoT reasoning in promoting understanding performance via long-chain thoughts. Wang \etal~\cite{wang2025mcot_survey} provided a comprehensive survey categorizing multimodal CoT methods. Despite strong performance, these explicit approaches incur significant computational costs due to lengthy text generation, often consuming thousands of tokens~\cite{xu2025towards}. This motivates the exploration of more efficient alternatives that preserve reasoning quality while reducing the generation burden.

\subsection{Latent Reasoning}

The limitations of text-based reasoning have motivated a growing body of work on latent reasoning that operates in continuous representation spaces~\cite{chen2025latent_survey}. CoCoNut~\cite{hao2024coconut} introduced a pioneering framework that trains language models to reason in a continuous latent space, replacing discrete text tokens with continuous thought vectors, and demonstrated that the continuous space can encode multiple alternative reasoning paths. CODI~\cite{shen2025codi} proposed compressing chain-of-thought into continuous representations via self-distillation. SoftCoT~\cite{xu2025softcot} introduced a small assistant model to generate continuous soft thought tokens, projected into the main LLM’s embedding space to boost reasoning accuracy and efficiency. CoLaR~\cite{tan2025colar} introduced dynamic latent compression that adaptively compresses reasoning chains based on problem difficulty. SIM-CoT~\cite{wei2026sim} used an auxiliary decoder to provide step-level implicit supervision, which stabilizes latent reasoning states to avoid collapse and enhances interpretability.

In the multi-modal setting, latent visual reasoning~\cite{li2025lvr,wang2026monet} has emerged as an approach that uses auxiliary images to generate latent thought tokens for reasoning. LVR~\cite{li2025lvr} performed implicit visual inference by requiring the latent visual representation to approximate the encoded features of auxiliary images. MoNet~\cite{wang2026monet} introduced a three-stage framework to gradually internalize auxiliary image features into the generation process of the backbone model. However, these approaches require auxiliary image annotations or generation models, resulting in high labelling costs and limited applicability to scenarios where visual intermediates are not readily available. Our proposed \ours{} differs fundamentally by operating entirely in the language model's latent space, requiring no auxiliary visual annotations while providing comprehensive supervision through forward, backward, and internal signals.

%=============================================

\section{Method}
\label{sec:method}

In this section, we present \ours{}, our framework for teaching multi-modal models to reason through latent thought representations. We first provide an overview of the framework (\cref{sec:overview}), then describe the latent thought representation (\cref{sec:latent_repr}), the external decoder supervision combining forward and backward decoding (\cref{sec:decoder_supervision}), the internal step-level supervision (\cref{sec:internal_supervision}), and the complete training objective (\cref{sec:training}).

\subsection{Overview}
\label{sec:overview}

Given a multi-modal input consisting of an image $\mathbf{v}$ and a text query $\mathbf{x}$, a standard MLLM first encodes the image through a vision encoder to obtain visual tokens $\mathbf{z}_v = \text{Encoder}(\mathbf{v})$, which are concatenated with the text tokens as input to the language model. In text-based CoT, the model generates explicit reasoning steps $\mathbf{r}_1, \mathbf{r}_2, \ldots, \mathbf{r}_L$ as natural language text, followed by the final answer $\mathbf{y}$:
\begin{equation}
    P(\mathbf{y} | \mathbf{v}, \mathbf{x}) = \sum_{\mathbf{r}_{1:L}} P(\mathbf{y} | \mathbf{r}_{1:L}, \mathbf{v}, \mathbf{x}) \prod_{l=1}^{L} P(\mathbf{r}_l | \mathbf{r}_{<l}, \mathbf{v}, \mathbf{x}).
    \label{eq:cot}
\end{equation}

\ours{} replaces the text reasoning steps with latent thought vectors $\mathbf{h}_1, \mathbf{h}_2, \ldots, \mathbf{h}_K \in \mathbb{R}^d$ with $K \ll L$, where $d$ denotes the hidden dimension of the language model. The model directly produces latent representations at designated reasoning positions, bypassing the text generation bottleneck while preserving multi-path exploration in the latent space:
\begin{equation}
    P(\mathbf{y} | \mathbf{v}, \mathbf{x}) = P(\mathbf{y} | \mathbf{h}_{1:K}, \mathbf{v}, \mathbf{x}),
    \label{eq:colt}
\end{equation}
where each $\mathbf{h}_k$ is obtained from the language model's hidden states at the corresponding latent thought position.

\begin{figure}[t]
    \centering
\centerline{\includegraphics[width=1.0\textwidth]{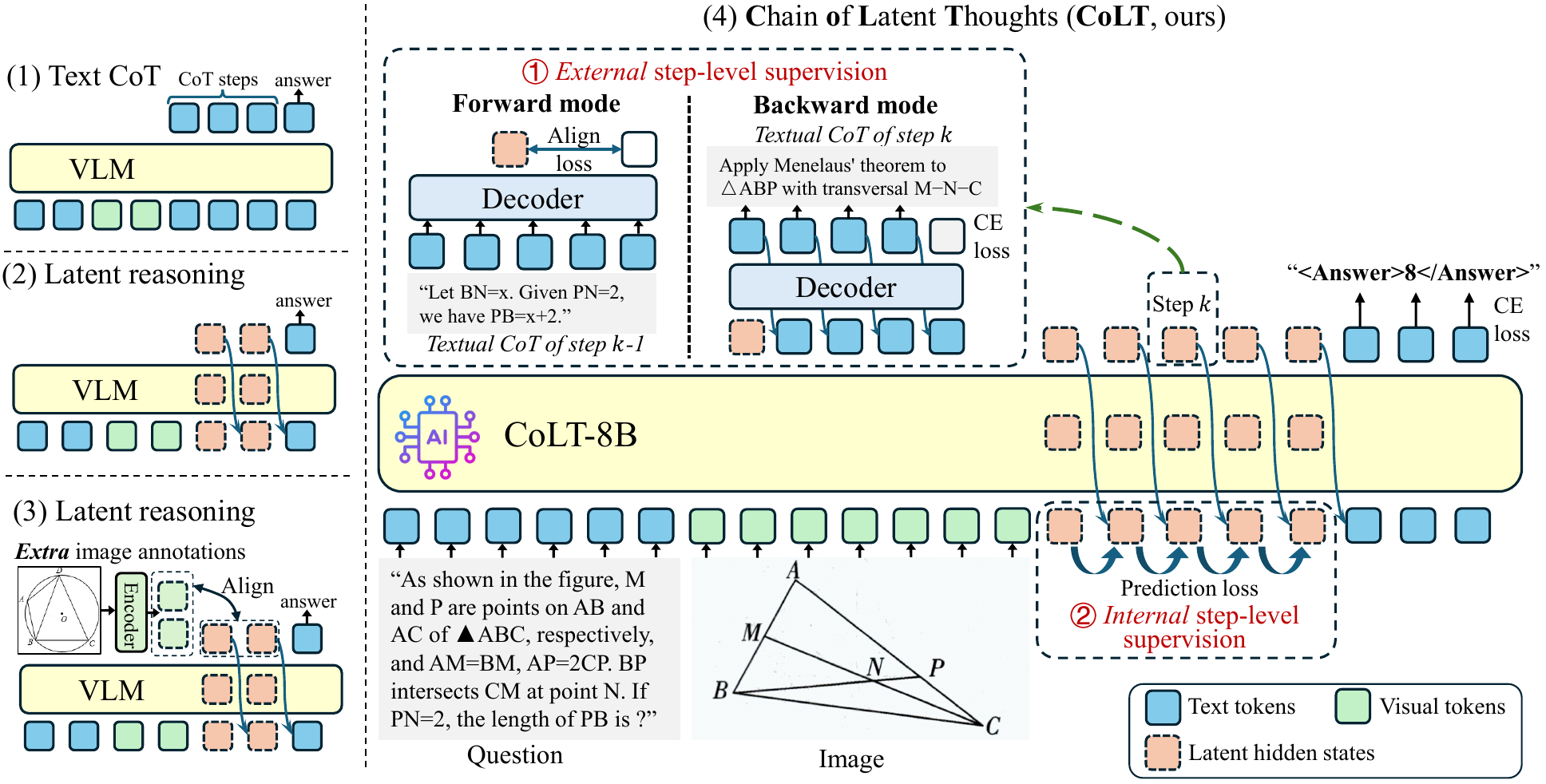}}
    \caption{Comparison of reasoning paradigms and overview of \ours{}. Left: (1)~text CoT generates verbose reasoning tokens before the answer; (2)~latent reasoning replaces text with latent states but lacks explicit supervision; (3)~latent visual reasoning aligns latent states with extra image annotations via an encoder. Right: \ours{} generates latent thought vectors $\mathbf{h}_1, \ldots, \mathbf{h}_K$ regulated by two complementary mechanisms: external step-level supervision (a decoder $D_\phi$ operating in forward and backward modes) and internal step-to-step prediction.}
\label{fig:overview}

\end{figure}

The key challenge in latent reasoning is ensuring that the latent thoughts encode meaningful and structured reasoning information, despite being unconstrained continuous vectors. The core contribution of \ours{} is to introduce fine-grained \emph{step-level supervision} that guides the latent thinking process. Specifically, we design two complementary supervision mechanisms:
\begin{itemize}
    \item \textbf{External decoder supervision} (\cref{sec:decoder_supervision}): We introduce a lightweight external decoder $D_\phi$, a smaller language model from the same model family as the backbone that shares the same tokenizer and vocabulary. The decoder provides step-level supervision from two directions: \emph{forward decoding} that decodes latent thoughts into the textual reasoning of the next step, and \emph{backward decoding} that aligns decoder hidden states with the model's latent thoughts given preceding textual context.
%A lightweight external decoder $D_\phi$, a smaller language model from the same family as the backbone, provides step-level supervision from two directions: \emph{forward decoding} that decodes latent thoughts into the textual reasoning of the next step, and \emph{backward decoding} that aligns decoder hidden states with the model's latent thoughts.
    \item \textbf{Internal step-level supervision} (\cref{sec:internal_supervision}): A lightweight projection head that encourages coherent transitions between consecutive latent steps by predicting the next step's representation from the current one.
\end{itemize}
Together, these signals comprehensively regulate the latent reasoning process, ensuring that each latent thought encodes progressively meaningful semantics. \Cref{fig:overview} compares \ours{} with existing reasoning paradigms (text CoT, latent reasoning, and latent visual reasoning) and illustrates our framework.

\subsection{Latent Thought Representation}
\label{sec:latent_repr}

To enable structured latent reasoning, we divide the reasoning process into $K$ discrete steps based on the logical structure of the chain-of-thought annotations. At each step, instead of generating explicit text tokens, the model directly produces a hidden state that serves as the latent thought vector. No special tokens are introduced into the vocabulary, and the latent thoughts exist purely as continuous hidden representations. The latent thought at step $k$ is obtained from the last hidden layer of the language model:
\begin{equation}
    \mathbf{h}_k = \text{LM}^{\text{last}}(\mathbf{z}_v, \mathbf{x}, \mathbf{h}_1, \ldots, \mathbf{h}_{k-1}) \in \mathbb{R}^d,
    \label{eq:latent_thought}
\end{equation}
where $\text{LM}^{\text{last}}(\cdot)$ extracts the last-layer hidden state at the current reasoning position. Each latent thought $\mathbf{h}_k$ is directly fed back into the language model as the input embedding for the next position, allowing subsequent latent thoughts and the final answer to attend to all preceding latent representations through the standard self-attention mechanism. This design preserves the autoregressive nature of the language model while enabling reasoning in the continuous latent space, avoiding any overhead from extending the model's vocabulary.

During training, we assume access to paired data where explicit text reasoning chains $\mathbf{r}_1, \ldots, \mathbf{r}_L$ are available (from existing CoT annotations or distilled from a teacher model). Each latent thought corresponds to $\frac{L}{K}$ consecutive textual tokens. The latent thought vectors are supervised to encode the same reasoning information through the mechanisms described below, and at inference time, the model uses only the latent thoughts without generating any text reasoning.

\subsection{External Decoder Supervision}
\label{sec:decoder_supervision}

The external decoder $D_\phi$ provides step-level supervision from two complementary directions: forward decoding that ensures latent thoughts encode sufficient information to produce textual reasoning, and backward decoding that anchors the latent space to the semantic structure of explicit reasoning chains.

\textbf{Forward decoding.} The forward decoding supervision ensures that the latent thoughts encode sufficient information to reconstruct the explicit reasoning process. The external decoder $D_\phi$ receives the preceding latent thought $\mathbf{h}_{k}$ as input and is trained to decode the textual reasoning content of the next step $\mathbf{r}_{k+1}$:
\begin{equation}
    \hat{\mathbf{r}}_{k+1} = D_\phi( \mathbf{h}_k),
    \label{eq:forward_decode}
\end{equation}
where $D_\phi$ autoregressively generates the text tokens conditioned on the preceding latent thoughts. Since $D_\phi$ is a smaller language model from the same model family and shares the same vocabulary as the main backbone, it can naturally decode the latent thoughts into coherent textual reasoning. The forward decoding loss is defined as the negative log-likelihood of the ground-truth textual reasoning:
\begin{equation}
    \Lfwd = -\frac{1}{K} \sum_{k=0}^{K-1} \sum_{t=1}^{|\mathbf{r}_{k+1}|} \log P_{D_\phi}(r_{k+1,t} | r_{k+1,<t}, \mathbf{h}_{k}),
    \label{eq:loss_fwd}
\end{equation}
where $r_{k+1,t}$ is the $t$-th token of the $(k+1)$-th reasoning step. This forward supervision provides a strong training signal: the latent thought at each step must capture enough information to enable accurate generation of the next reasoning step. The gradients from the decoder flow back through the latent thoughts to the main language model, encouraging the formation of informative latent representations.

\textbf{Backward decoding.} While forward decoding maps from latent to text, backward decoding provides the reverse constraint by anchoring the latent space to the semantic structure of text reasoning. The decoder $D_\phi$ receives the preceding textual reasoning steps $\mathbf{r}_{k-1}$ as input and produces a hidden state $\tilde{\mathbf{h}}_k$ trained to align with the model's latent thought:
\begin{equation}
    \tilde{\mathbf{h}}_k = D_\phi^{\text{last}}(\mathbf{r}_{k-1}),
    \label{eq:backward_encode}
\end{equation}
where $D_\phi^{\text{last}}$ extracts the last-layer hidden state of the decoder after processing the text input. Since $D_\phi$ belongs to the same model family and shares the same vocabulary as the main backbone, the decoder's text-conditioned hidden states naturally reside in a compatible representation space, enabling effective alignment without additional projection layers. The backward decoding loss enforces alignment between $\tilde{\mathbf{h}}_k$ and the corresponding latent thought $\mathbf{h}_k$:
\begin{equation}
    \Lbwd = \frac{1}{K} \sum_{k=1}^{K} \left\| \frac{\tilde{\mathbf{h}}_k}{\|\tilde{\mathbf{h}}_k\|} - \frac{\text{sg}(\mathbf{h}_k)}{\|\text{sg}(\mathbf{h}_k)\|} \right\|^2,
    \label{eq:loss_bwd}
\end{equation}
where $\text{sg}(\cdot)$ denotes the stop-gradient operator applied to prevent the latent thoughts from collapsing to match trivial decoder outputs. The normalization ensures that alignment is measured in direction rather than magnitude, providing more stable training dynamics. Together, forward and backward signals provide comprehensive decoder-based regulation: forward decoding ensures the latent thoughts \emph{can produce} reasoning text, while backward decoding ensures they \emph{align with} the semantic content of reasoning steps.

\subsection{Internal Step-Level Supervision}
\label{sec:internal_supervision}

Beyond the external decoder supervision, we introduce an internal supervision mechanism that directly regularizes the transitions between consecutive latent thought steps. The key intuition is that adjacent latent steps should have a predictable relationship, mirroring the logical progression in explicit reasoning chains. We implement this through a lightweight projection head $f_\theta: \mathbb{R}^d \to \mathbb{R}^d$ (a two-layer MLP with GELU activation) that predicts the next step's latent thought from the current one:
\begin{equation}
    \hat{\mathbf{h}}_{k+1} = f_\theta(\mathbf{h}_k),
    \label{eq:internal_pred}
\end{equation}
with the internal supervision loss:
\begin{equation}
    \Lint = \frac{1}{K-1} \sum_{k=1}^{K-1} \left(1 - \frac{\hat{\mathbf{h}}_{k+1} \cdot \text{sg}(\mathbf{h}_{k+1})}{\|\hat{\mathbf{h}}_{k+1}\| \cdot \|\text{sg}(\mathbf{h}_{k+1})\|}\right),
    \label{eq:loss_int}
\end{equation}
where we use cosine similarity with a stop-gradient on the target to prevent representational collapse. The internal supervision complements the external decoder signals: while the decoder-based losses anchor the latent space to the \emph{content} of reasoning through text-latent correspondence, the internal prediction loss regularizes the \emph{structure} of the latent chain through step-to-step consistency.

\subsection{Training Objective}
\label{sec:training}

The complete training objective of \ours{} combines the standard task loss with the three supervision signals:
\begin{equation}
    \mathcal{L} = \Ltask + \alpha \Lfwd + \beta \Lbwd + \gamma \Lint,
    \label{eq:total_loss}
\end{equation}
where $\Ltask$ is the standard next-token prediction loss for generating the final answer $\mathbf{y}$ conditioned on the latent thoughts:
\begin{equation}
    \Ltask = -\sum_{t=1}^{|\mathbf{y}|} \log P_\text{LM}(y_t | y_{<t}, \mathbf{h}_{1:K}, \mathbf{v}, \mathbf{x}),
    \label{eq:loss_task}
\end{equation}
and $\alpha$, $\beta$, $\gamma$ are hyperparameters controlling the relative importance of each supervision signal.

\textbf{Training procedure.} The training of \ours{} is mainly composed of \emph{supervised fine-tuning} (SFT). Specifically, we train the external decoder and LLM of the backbone with the combined objective in \cref{eq:total_loss} on CoT annotations from the OneThinker dataset~\cite{feng2025onethinker}, which provides high-quality textual reasoning annotations across diverse visual tasks. We use only the image subset, which covers multiple complex visual reasoning tasks including rule-based question answering, open-ended question answering, visual captioning, spatial grounding, temporal grounding, spatio-temporal grounding, tracking and so on. The annotations of each training sample is annotated by a pair of CoT chains wrapped by $\langle\text{Think}\rangle$ and $\langle\text{Answer}\rangle$ tokens, which naturally enable us to divide the training data into CoT steps and answer parts. This stage teaches the model to produce structured latent thoughts regulated by the three supervision signals. During inference, we discard the external decoder as well as the losses and directly ask the model to output $K$ latent vectors followed by answer generation, enabling great efficiency.

%In the second stage (\emph{reinforcement learning}), we apply Group Relative Policy Optimization (GRPO)~\cite{shao2024deepseekmath} on 50K CoT samples, allowing the latent thoughts to update freely without the decoder and internal supervision constraints. By removing the auxiliary losses and optimizing solely through outcome-based rewards, the model is encouraged to explore latent reasoning strategies beyond what the supervised signals prescribe. During inference, the external decoder and projection head are discarded entirely, incurring no additional computational overhead.

%=============================================
\section{Experiments}
\label{sec:experiments}

\subsection{Experimental Setup}
\label{sec:setup}

\textbf{Benchmarks.}
We evaluate \ours{} on eight diverse multi-modal benchmarks: SeedBench~\cite{li2024seed} for comprehensive visual understanding and reasoning; MMBench~\cite{liu2024mmbench} covering multiple visual perception and reasoning abilities; ChartQA~\cite{masry2022chartqa} for chart comprehension and data reasoning; TextVQA~\cite{singh2019towards} targeting text-rich visual question answering; ScienceQA~\cite{lu2022scienceqa} for multi-modal science question answering; MMStar~\cite{chen2024mmstar} with carefully curated vision-indispensable problems; AI2D~\cite{kembhavi2016ai2d} focusing on diagram understanding; and MMT-Bench~\cite{ying2024mmtbench} for comprehensive multitask evaluation.

\textbf{Implementation details.}
We build \ours{} on top of Qwen3-VL-8B-Instruct~\cite{bai2025qwen3}, and instantiate the external decoder $D_\phi$ as Qwen3-0.6B~\cite{yang2025qwen3}, a smaller model from the same Qwen3 family that shares the same tokenizer and vocabulary. The projection head $f_\theta$ is a 2-layer MLP with GELU activation. We set the number of latent thought steps $K=3$. The loss weights are $\alpha=0.2$, $\beta=0.2$, and $\gamma=0.2$. We dynamically split the textual CoT annotations into $K$ splits during training to enhance the ability of model understanding. We train all models on the image subset of Onethinker dataset~\cite{feng2026onethinker} with one epoch. All models are evaluated with the same prompt and same configurations to ensure fairness. We use the AdamW~\cite{loshchilov2019adamw} optimizer with a cosine learning rate schedule and a batch size of 8. More details can be referred to our codebase.

\subsection{Main Results}
\label{sec:main_results}

\Cref{tab:comparison} presents a comprehensive comparison of \ours{} against proprietary models, open-source MLLMs, latent reasoning methods, and latent visual reasoning approaches across all eight benchmarks.

\textbf{Baseline comparison.} Compared to direct answer (no reasoning), \ours{} improves average performance by +9.6\% (79.1 vs.\ 69.5), confirming that latent thoughts provide substantial reasoning capacity despite generating only 3 continuous vectors. Compared to textual reasoning with explicit chain-of-thought, \ours{} achieves +3.4\% higher average (79.1 vs.\ 75.7), with the largest gains on ChartQA (+9.6\%) and TextVQA (+6.1\%), while using significantly fewer tokens.

\begin{table}[t]
\caption{Comprehensive comparison across proprietary models, open-source MLLMs, latent reasoning methods and latent visual reasoning methods. 'SQA` is the abbreviation of `ScienceQA' benchmark. $^\dagger$: methods built on the Qwen3-VL-8B backbone. Best non-proprietary results in \textbf{bold}.}
\label{tab:comparison}
\centering
\setlength{\tabcolsep}{3pt}
\small
\resizebox{\textwidth}{!}{
\begin{tabular}{l|c|cccccccc|c}
\toprule
\textbf{Method} & \textbf{Size} & \textbf{SeedBench} & \textbf{MMBench} & \textbf{ChartQA} & \textbf{TextVQA} & \textbf{SQA} & \textbf{MMStar} & \textbf{AI2D} & \textbf{MMT} & \textbf{Avg.} \\
\midrule
\multicolumn{11}{l}{\emph{Proprietary Models}} \\
\rowcolor{Gray} GPT-5-high & - & - & 83.8 & 86.2 & - & 97.4 & 76.4 & 89.7 & 77.2 & - \\
\rowcolor{Gray} Claude-Opus-4.1 & - & - & 83.0 & 83.3 & - & - & 71.0 & 84.4 & - & - \\
\rowcolor{Gray} Gemini-2.5-Pro & - & - & 90.1 & 59.7 & - & 96.2 & 77.5 & 88.4 & 75.4 & - \\
\midrule
\multicolumn{11}{l}{\textit{Open-Source MLLMs (7-8B scale)}} \\
LLaVA-OneVision-1.5 & 8B & 77.3 & 84.1 & 86.5 & - & 95.0 & 67.7 & 84.2 & - & - \\
InternVL3 & 8B & - & 83.4 & 86.6 & 80.2  & - & 68.2 & 85.2 & 65.0 & - \\
%Mini-Gemini             & 7B  & 38.2 & 69.3 & 28.5 & 46.2 & 82.5 & 44.9 & 66.0 & 53.0 & 53.6 \\
\midrule
\multicolumn{11}{l}{\emph{Latent Reasoning$^\dagger$}} \\
CCoT  & 8B & 63.4 & 78.4 & 60.4 & 63.1 & 87.0 & 60.3 & 77.0 & 57.1 & 68.3 \\
ICoT  & 8B & 64.7 & 79.1 & 62.1 & 65.6 & 87.6 & 61.2 & 77.8 & 57.9 & 69.5 \\
CODI  & 8B & 68.6 & 80.2 & 61.6 & 67.2 & 88.4 & 62.5 & 79.0 & 59.1 & 70.8 \\
SoftCoT & 8B & 69.2 & 81.0 & 64.5 & 68.4 & 89.5 & 64.0 & 80.4 & 60.6 & 72.2 \\
SIM-CoT & 8B & 72.4 & 82.0 & 66.7 & 70.2 & 90.8 & 65.6 & 82.0 & 62.4 & 74.0 \\
\midrule
\multicolumn{11}{l}{\emph{Latent Visual Reasoning}} \\
Multimodal CoT$^\dagger$ & 8B & 63.2 & 79.5 & 62.1 & 66.1 & 87.8 & 61.5 & 78.2 & 58.3 & 69.6 \\
LaCoT & 7B & 67.4 & - & 64.6 & 67.6 & - & - & - & - & - \\
LVR  & 7B & 71.2 & 81.5 & 67.2 & 70.2 & 90.2 & 64.8 & 81.3 & 62.6 & 73.6 \\
\midrule

Qwen3-VL (Direct answer) & 8B & 66.2 & 76.2 & 64.1 & 74.3 & 85.3 & 58.6 & 75.3 & 55.6 & 69.5 \\
Qwen3-VL (Textual reasoning)& 8B & 76.4 & 83.4 & 65.1 & 75.2  & 91.8 & 67.1 & 83.6 & 63.3 & 75.7 \\
\rowcolor{bestcolor}
\textbf{\ours{}$^\dagger$} & 8B & \textbf{77.5} & \textbf{84.6} & \textbf{74.7} & \textbf{81.3} & \textbf{92.8} & \textbf{68.9} & \textbf{85.4} & \textbf{67.4} & \textbf{79.1} \\
\scriptsize
 Relative improvement &  & \textcolor{seagreen}{\textbf{+1.1}} & \textcolor{seagreen}{\textbf{+1.2}} & \textcolor{seagreen}{\textbf{+9.6}} & \textcolor{seagreen}{\textbf{+6.1}} & \textcolor{seagreen}{\textbf{+1.0}} & \textcolor{seagreen}{\textbf{+1.8}} & \textcolor{seagreen}{\textbf{+1.8}} & \textcolor{seagreen}{\textbf{+4.1}} & \textcolor{seagreen}{\textbf{+3.3}} \\
% &  & \textcolor{seagreen}{+5.2} & \textcolor{seagreen}{+1.2} & \textcolor{seagreen}{+1.8} & \textcolor{seagreen}{+2.0} & \textcolor{seagreen}{+1.7} & \textcolor{seagreen}{+1.8} & \textcolor{seagreen}{+1.2} & \textcolor{seagreen}{+1.6} & \textcolor{seagreen}{\textbf{+2.0}} \\
\bottomrule
\end{tabular}
}
\end{table}

\textbf{Comparison with latent reasoning methods.} We compare against five latent reasoning methods, all built on the same Qwen3-VL-8B backbone for fair comparison. CCoT~\cite{cheng2024compressed} compresses chain-of-thought into dense representations; ICoT~\cite{deng2023implicit} internalizes reasoning via distillation; CODI~\cite{shen2025codi} distills CoT into latent thoughts; SoftCoT~\cite{xu2025softcot} uses soft continuous embeddings; and SIM-CoT~\cite{wei2026sim} aligns implicit tokens with explicit reasoning. \ours{} surpasses the strongest baseline SIM-CoT by +5.1\% on average (79.1 vs.\ 74.0), with consistent gains including +5.1\% on SeedBench, +8.0\% on ChartQA, and +11.1\% on TextVQA, demonstrating the effectiveness of three-way step-level supervision.

\textbf{Comparison with latent visual reasoning.} We further compare with approaches that incorporate auxiliary images for additional supervision: Multimodal CoT~\cite{zhang2023multimodal} decomposes reasoning into rationale generation and answer inference; LaCoT~\cite{sun2025lacot} performs visual reasoning through continuous latent representations; and LVR~\cite{li2025lvr} generates latent visual tokens as intermediate representations. \ours{} outperforms the strongest baseline LVR by +5.5\% on average (79.1 vs.\ 73.6), with notable gains on TextVQA (+11.1\%) and SeedBench (+6.3\%) while eliminating the need for auxiliary image annotations, validating our design of operating in the language model's latent space.

\subsection{Ablation Study}
\label{sec:ablation}

We conduct comprehensive ablation studies to analyze the contribution of each component in \ours{}. All ablations use the Qwen3-VL-8B backbone and report results on four representative benchmarks.

\textbf{Effect of supervision signals.}
\Cref{tab:ablation_loss} presents a systematic ablation over all combinations of the three supervision signals. Several key findings emerge: (1)~Among individual losses, backward decoding alone achieves the highest single-component accuracy ($69.3\%$), outperforming forward decoding ($68.0\%$) and internal supervision ($65.8\%$). (2)~Every pairwise combination substantially improves over its constituent single losses, with $\Lfwd + \Lbwd$ ($71.9\%$) being the strongest pair. (3)~Adding the third signal to any pair consistently yields further gains, with backward decoding contributing the largest marginal improvement ($+4.6\%$ when added to $\Lfwd + \Lint$), followed by forward ($+3.5\%$) and internal ($+2.9\%$). Overall, this confirms all three losses provide non-redundant regularization and complete three-way supervision is essential for optimal performance.

\begin{table}[t]
\caption{Ablation study on supervision signal combinations. \checkmark{} indicates the corresponding loss is active during training.}
\setlength{\tabcolsep}{3pt}
\label{tab:ablation_loss}
\centering
\small
\begin{tabular}{ccc|cccc|c}
\toprule
$\Lfwd$ & $\Lbwd$ & $\Lint$ & \textbf{SeedBench} & \textbf{MMStar} & \textbf{AI2D} & \textbf{MMT} & \textbf{Avg.} \\
\midrule
 &  &  & 65.8 & 57.4 & 74.8 & 56.2 & 63.6 \\
\checkmark &  &  & 70.5 & 61.9 & 79.2 & 60.2 & 68.0 \\
 & \checkmark &  & 71.8 & 63.2 & 80.4 & 61.9 & 69.3 \\
 &  & \checkmark & 68.1 & 59.6 & 77.1 & 58.4 & 65.8 \\
\midrule
\checkmark & \checkmark &  & 74.3 & 65.9 & 83.1 & 64.4 & 71.9 \\
\checkmark &  & \checkmark & 72.8 & 63.9 & 81.6 & 62.6 & 70.2 \\
 & \checkmark & \checkmark & 73.5 & 65.2 & 82.6 & 63.9 & 71.3 \\
\midrule
\rowcolor{bestcolor}
\checkmark & \checkmark & \checkmark & \textbf{77.5} & \textbf{68.9} & \textbf{85.4} & \textbf{67.4} & \textbf{74.8} \\
\bottomrule
\end{tabular}
\end{table}

\textbf{Effect of decoder architecture.}
We investigate the impact of the external decoder size in \cref{tab:ablation_decoder}. All decoder variants are from the Qwen3 family and share the same tokenizer and vocabulary. Scaling the decoder from Qwen3-0.6B to Qwen3-8B yields only marginal improvements: Qwen3-0.6B performs only slightly worse than Qwen3-1.7B ($-$0.3\%), and Qwen3-1.7B nearly matches Qwen3-4B ($-$0.2\%). Further scaling to Qwen3-8B provides a negligible +0.1\% gain. These results demonstrate that even a lightweight 0.6B decoder provides effective step-level supervision, and we use Qwen3-0.6B as the default for its best accuracy-efficiency trade-off.

\begin{table}[t]
\caption{Ablation on external decoder architecture. All decoders are from the Qwen3 family and share the same vocabulary as the backbone.}
\label{tab:ablation_decoder}
\centering
\small
\setlength{\tabcolsep}{3pt}
\begin{tabular}{l|c|cccc|c}
\toprule
\textbf{Decoder} & \textbf{Params} & \textbf{SeedBench} & \textbf{MMStar} & \textbf{AI2D} & \textbf{MMT} & \textbf{Avg.} \\
\midrule
\rowcolor{bestcolor}
\textbf{Qwen3-0.6B}      & 0.6B  & \textbf{77.5} & \textbf{68.9} & \textbf{85.4} & \textbf{67.4} & \textbf{74.8} \\
Qwen3-1.7B      & 1.7B  & 77.8 & 69.2 & 85.7 & 67.7 & 75.1 \\
Qwen3-4B        & 4.0B  & 78.0 & 69.4 & 85.9 & 67.9 & 75.3 \\
Qwen3-8B        & 8.0B  & 78.1 & 69.5 & 86.0 & 68.0 & 75.4 \\
\bottomrule
\end{tabular}
\end{table}

\begin{table}[t]
\caption{Effect of the number of latent thought steps $K$ on four benchmarks. Performance peaks at $K{=}3$ and gradually decreases with larger $K$.}

\label{tab:num_steps}
\centering
\small
\setlength{\tabcolsep}{3pt}
\begin{tabular}{c|cccc|c}
\toprule
$K$ & \textbf{SeedBench} & \textbf{MMStar} & \textbf{AI2D} & \textbf{MMT} & \textbf{Avg.} \\
\midrule
1  & 71.8 & 62.9 & 79.8 & 61.2 & 68.9 \\
2  & 75.3 & 66.4 & 83.4 & 64.9 & 72.5 \\
\rowcolor{bestcolor}
3  & \textbf{77.5} & \textbf{68.9} & \textbf{85.4} & \textbf{67.4} & \textbf{74.8} \\
4  & 77.1 & 68.4 & 85.1 & 66.9 & 74.4 \\
6  & 76.6 & 67.9 & 84.6 & 66.4 & 73.9 \\
8  & 76.1 & 67.4 & 84.2 & 65.9 & 73.4 \\
\bottomrule
\end{tabular}
\end{table}

\textbf{Effect of number of latent steps.}
\Cref{tab:num_steps} presents the performance of \ours{} as a function of the number of latent thought steps $K$. Performance increases rapidly from $K{=}1$ (68.9\%) through $K{=}2$ (72.5\%) to $K{=}3$ (74.8\%), beyond which it gradually decreases: $K{=}4$ (74.4\%), $K{=}6$ (73.9\%), and $K{=}8$ (73.4\%). This suggests that $K{=}3$ provides the optimal trade-off, where too few steps limit reasoning capacity while excessive steps introduce redundancy that slightly degrades performance. We use $K{=}3$ as the default.

\begin{table}[t]
\caption{Cross-$K$ generalization: performance (\%) on individual benchmarks when training and testing with different numbers of latent steps. Best result in \textbf{bold}.}
\label{tab:cross_k}
\centering
\setlength{\baselineskip}{1.1\baselineskip}
\setlength{\tabcolsep}{1pt}
\small
\begin{minipage}[t]{0.31\linewidth}
\centering
\textbf{(a) SeedBench}\\[2pt]
\begin{tabular}{c|ccccc}
\toprule
\multirow{2}{*}{\textbf{\makecell{Train \\K}}} & \multicolumn{4}{c}{\textbf{Test $K$}} \\
 & $1$ & $3$ & $6$ & $8$ \\
\midrule
$1$  & \cellcolor{bestcolor}\textbf{71.8} & 71.1 & 70.3 & 69.8 \\
$3$  & 72.3 & \cellcolor{bestcolor}\textbf{77.5} & 76.8 & 76.3 \\
$6$  & 71.3 & 76.3 & \cellcolor{bestcolor}\textbf{76.6} & 76.3 \\
$8$  & 70.8 & 75.8 & 75.9 & \cellcolor{bestcolor}\textbf{76.1} \\
\bottomrule
\end{tabular}
\end{minipage}
\hfill
\begin{minipage}[t]{0.31\linewidth}
\centering
\textbf{(b) MMStar}\\[1pt]
\begin{tabular}{c|ccccc}
\toprule
\multirow{2}{*}{\textbf{\makecell{Train \\K}}} & \multicolumn{4}{c}{\textbf{Test $K$}} \\
 & $1$ & $3$ & $6$ & $8$ \\
\midrule
$1$  & \cellcolor{bestcolor}\textbf{62.9} & 62.2 & 61.6 & 61.2 \\
$3$  & 63.4 & \cellcolor{bestcolor}\textbf{68.9} & 68.2 & 67.7 \\
$6$  & 62.4 & 67.6 & \cellcolor{bestcolor}\textbf{67.9} & 67.6 \\
$8$  & 61.9 & 67.2 & 67.2 & \cellcolor{bestcolor}\textbf{67.4} \\
\bottomrule
\end{tabular}
\end{minipage}
\hfill
\begin{minipage}[t]{0.31\linewidth}
\centering
\textbf{(c) MMT}\\[1pt]
\begin{tabular}{c|ccccc}
\toprule
\multirow{2}{*}{\textbf{\makecell{Train \\K}}} & \multicolumn{4}{c}{\textbf{Test $K$}} \\
 & $1$ & $3$ & $6$ & $8$ \\
\midrule
$1$  & \cellcolor{bestcolor}\textbf{61.2} & 60.4 & 59.9 & 59.4 \\
$3$  & 61.9 & \cellcolor{bestcolor}\textbf{67.4} & 66.7 & 66.2 \\
$6$  & 60.9 & 66.2 & \cellcolor{bestcolor}\textbf{66.4} & 66.2 \\
$8$  & 60.4 & 65.6 & 65.7 & \cellcolor{bestcolor}\textbf{65.9} \\
\bottomrule
\end{tabular}
\end{minipage}
\hfill
% \begin{minipage}[t]{0.235\linewidth}
% \centering
% \textbf{(d) ScienceQA}\\[1pt]
% \begin{tabular}{c|ccccc}
% \toprule
% & \multicolumn{4}{c}{\textbf{Test $K$}} \\
% \textbf{Train} & $4$ & $8$ & $12$ & $16$ \\
% \midrule
% $4$  & \cellcolor{bestcolor}\textbf{92.3} & 91.8 & 91.4 & 91.0 \\
% $8$  & 92.7 & \cellcolor{bestcolor}\textbf{93.1} & 92.8 & 92.5 \\
% $12$ & 92.2 & 92.9 & \cellcolor{bestcolor}\textbf{93.1} & 92.8 \\
% $16$ & 91.8 & 92.6 & 92.9 & \cellcolor{bestcolor}\textbf{93.0} \\
% \bottomrule
% \end{tabular}
% \end{minipage}

\end{table}

\textbf{Generalization across latent steps.}
Beyond the default setting where training and testing use the same $K$, we investigate cross-$K$ generalization in \cref{tab:cross_k}. The diagonal entries (matching $K$) consistently achieve the best performance, confirming that the model learns step-specific reasoning patterns. Notably, \ours{} exhibits strong robustness to $K$ mismatch: models trained with $K{=}3$ and tested at $K{=}6$ incur only minor drops (0.7\% on SeedBench, 0.7\% on MMStar, 0.7\% on MMT), and testing at $K{=}8$ similarly retains most of the performance with drops of merely 1.2\%. Even the largest mismatch (training with $K{=}1$, testing at $K{=}8$) causes at most 2.0\% degradation, confirming $K{=}3$ as a practical default that generalizes well across different test-time step counts.

\subsection{Robustness to Input Noise}

Compared to explicit text decoding, latent reasoning can preserve the probability of multi-path reasoning within latent spaces and thus can perform more robustly to input noise. To assess the robustness of latent reasoning under noisy inputs, we evaluate \ours{} against Direct Answer and Text CoT under two categories of perturbations. For visual noise, we apply Gaussian blur ($\sigma \in \{1, 2, 3\}$) and random occlusion (masking 10\%/20\%/30\% of image patches). For text noise, we introduce character-level spelling errors (5\%/10\% of tokens) and key information deletion (10\%/20\% of question tokens). \Cref{tab:noise} reports per-benchmark accuracy degradation on the SeedBench, MMStar, and MMT benchmarks.

\begin{table}[t]
\caption{Robustness to input noise. Per-benchmark accuracy degradation (\%) under visual noise and text noise on SeedBench, MMStar, and MMT.}

\label{tab:noise}
\centering
\setlength{\tabcolsep}{2pt}
\small
\resizebox{0.9\textwidth}{!}{
\begin{tabular}{ll|ccc|ccc|ccc}
\toprule
\multirow{2}{*}{\textbf{Noise}} & \multirow{2}{*}{\textbf{Level}} & \multicolumn{3}{c|}{\textbf{SeedBench}} & \multicolumn{3}{c|}{\textbf{MMStar}} & \multicolumn{3}{c}{\textbf{MMT}} \\
& & \makecell{Direct \\Answer} & \makecell{Text \\CoT} & \ours{} & \makecell{Direct \\Answer}  & \makecell{Text \\CoT} & \ours{} & \makecell{Direct \\Answer}  & \makecell{Text \\CoT} & \ours{} \\
\midrule
\multicolumn{11}{l}{\emph{Visual Noise}} \\
Gauss.\ blur & $\sigma{=}1$ & $-$7.5 & $-$5.2 & \textbf{$-$2.8} & $-$8.8 & $-$6.0 & \textbf{$-$3.4} & $-$8.2 & $-$5.6 & \textbf{$-$3.0} \\
            & $\sigma{=}2$ & $-$11.8 & $-$8.0 & \textbf{$-$4.8} & $-$13.0 & $-$9.0 & \textbf{$-$5.5} & $-$12.2 & $-$8.5 & \textbf{$-$5.0} \\
            & $\sigma{=}3$ & $-$15.8 & $-$11.0 & \textbf{$-$6.8} & $-$17.5 & $-$12.5 & \textbf{$-$8.0} & $-$16.2 & $-$11.5 & \textbf{$-$7.2} \\
Occlusion & 10\% & $-$8.2 & $-$5.6 & \textbf{$-$3.0} & $-$9.2 & $-$6.4 & \textbf{$-$3.6} & $-$8.5 & $-$5.8 & \textbf{$-$3.2} \\
          & 20\% & $-$12.5 & $-$8.5 & \textbf{$-$5.2} & $-$14.0 & $-$9.8 & \textbf{$-$6.0} & $-$13.0 & $-$9.0 & \textbf{$-$5.5} \\
          & 30\% & $-$17.0 & $-$12.2 & \textbf{$-$7.8} & $-$19.0 & $-$13.8 & \textbf{$-$9.0} & $-$17.5 & $-$12.5 & \textbf{$-$8.2} \\
\midrule
\multicolumn{11}{l}{\emph{Text Noise}} \\
Spell.\ err. & 5\% & $-$7.8 & $-$5.5 & \textbf{$-$3.2} & $-$7.2 & $-$5.0 & \textbf{$-$2.8} & $-$8.0 & $-$5.8 & \textbf{$-$3.4} \\
            & 10\% & $-$12.2 & $-$8.2 & \textbf{$-$5.0} & $-$11.5 & $-$7.8 & \textbf{$-$4.5} & $-$12.5 & $-$8.5 & \textbf{$-$5.2} \\
Key del. & 10\% & $-$10.5 & $-$7.5 & \textbf{$-$4.2} & $-$9.5 & $-$6.8 & \textbf{$-$3.8} & $-$10.8 & $-$7.8 & \textbf{$-$4.5} \\
         & 20\% & $-$15.5 & $-$11.2 & \textbf{$-$6.8} & $-$14.2 & $-$10.0 & \textbf{$-$6.2} & $-$16.0 & $-$11.5 & \textbf{$-$7.2} \\
\bottomrule
\end{tabular}
}

\end{table}

\begin{table}[t]
\caption{Inference speed comparison on MMStar and SeedBench (per-sample average). All methods use the same Qwen3-VL-8B backbone on a single H200 GPU. Time is decomposed into input encoding and token generation (in seconds).}

\label{tab:speed}
\centering
\small
\begin{tabular}{l|c|cc|c|c|cc|c}
\toprule
\multirow{3}{*}{\textbf{Method}} & \multicolumn{4}{c|}{\textbf{MMStar}} & \multicolumn{4}{c}{\textbf{MMT-Bench}} \\
& \multirow{2}{*}{\makecell{CoT \\Tokens}} & \multicolumn{2}{c|}{Time} &  \multirow{2}{*}{Acc.} & \multirow{2}{*}{\makecell{CoT \\Tokens}} & \multicolumn{2}{c|}{Time} &  \multirow{2}{*}{Acc.} \\
& & Encode & Generate & & & Encode & Generate &  \\
\midrule
Direct answer    & 0         & 0.45 & 0.14 & 58.6 & 0         & 0.46 & 0.15 & 55.6 \\
Text CoT         & 142.1 & 0.47 & 7.24 & 67.1 & 138.5 & 0.48 & 7.38 & 63.3 \\
\textbf{\ours{}} & 3  & 0.44 &  0.32  & \textbf{68.9} &  3         & 0.45 & 0.33  & \textbf{67.4} \\
& (\scriptsize{\textbf{\textcolor{seagreen}{$\downarrow$ 47.4$\times$}}}) & & (\scriptsize{\textbf{\textcolor{seagreen}{$\downarrow$ 22.6$\times$}}}) & & (\scriptsize{\textbf{\textcolor{seagreen}{$\downarrow$ 46.2$\times$}}}) & & (\scriptsize{\textbf{\textcolor{seagreen}{$\downarrow$ 22.4$\times$}}}) & \\ 
\bottomrule
\end{tabular}

\end{table}

\ours{} consistently exhibits the lowest performance degradation across all noise conditions and benchmarks. Under visual noise, MMStar shows the highest sensitivity due to its reliance on vision-indispensable problems: at the most severe occlusion level (30\%), Direct Answer degrades by 19.0\%, Text CoT by 13.8\%, and \ours{} by only 9.0\%. In contrast, SeedBench is comparatively more resilient, with \ours{} degrading by at most 7.8\%. For text noise, MMT is the most affected due to its diverse multitask evaluation, while MMStar shows the smallest drops given its focus on vision-critical problems. Overall, \ours{}'s degradation ranges from 2.8\% to 9.0\% across all conditions, substantially lower than Direct Answer (7.2\% to 19.0\%) and Text CoT (5.0\% to 13.8\%). We attribute this robustness to the continuous nature of latent representations: whereas discrete text tokens propagate local input errors through the sequential decoding process, continuous latent vectors distribute information across all dimensions, enabling more graceful degradation under noise. Furthermore, the three-way supervision in \ours{} encourages latent states to capture abstract reasoning patterns rather than surface-level input features, making them inherently less sensitive to input perturbations.

\subsection{Inference Speed Comparison}
A key advantage of latent reasoning over text-based CoT is the reduced inference time. \Cref{tab:speed} decomposes the per-sample inference time on MMStar and MMT-Bench into input encoding and token generation. The encoding phase is roughly constant across methods (${\sim}$0.44--0.48\,s), as it processes the same inputs regardless of reasoning strategy. The critical difference lies in generation: text CoT requires 7.24\,s (MMStar) and 7.38\,s (MMT-Bench) to produce ${\sim}$142 and ${\sim}$139 reasoning tokens respectively, whereas \ours{} completes generation in 0.32\,s and 0.33\,s using only 3 latent vectors, achieving $22.6{\times}$ and $22.4{\times}$ reduction in generation time. Overall, \ours{} delivers $10.1{\times}$ (MMStar) and $10.1{\times}$ (MMT-Bench) end-to-end speedup while maintaining higher accuracy.

\begin{figure}[t]
    \centering
\centerline{\includegraphics[width=1.0\textwidth]{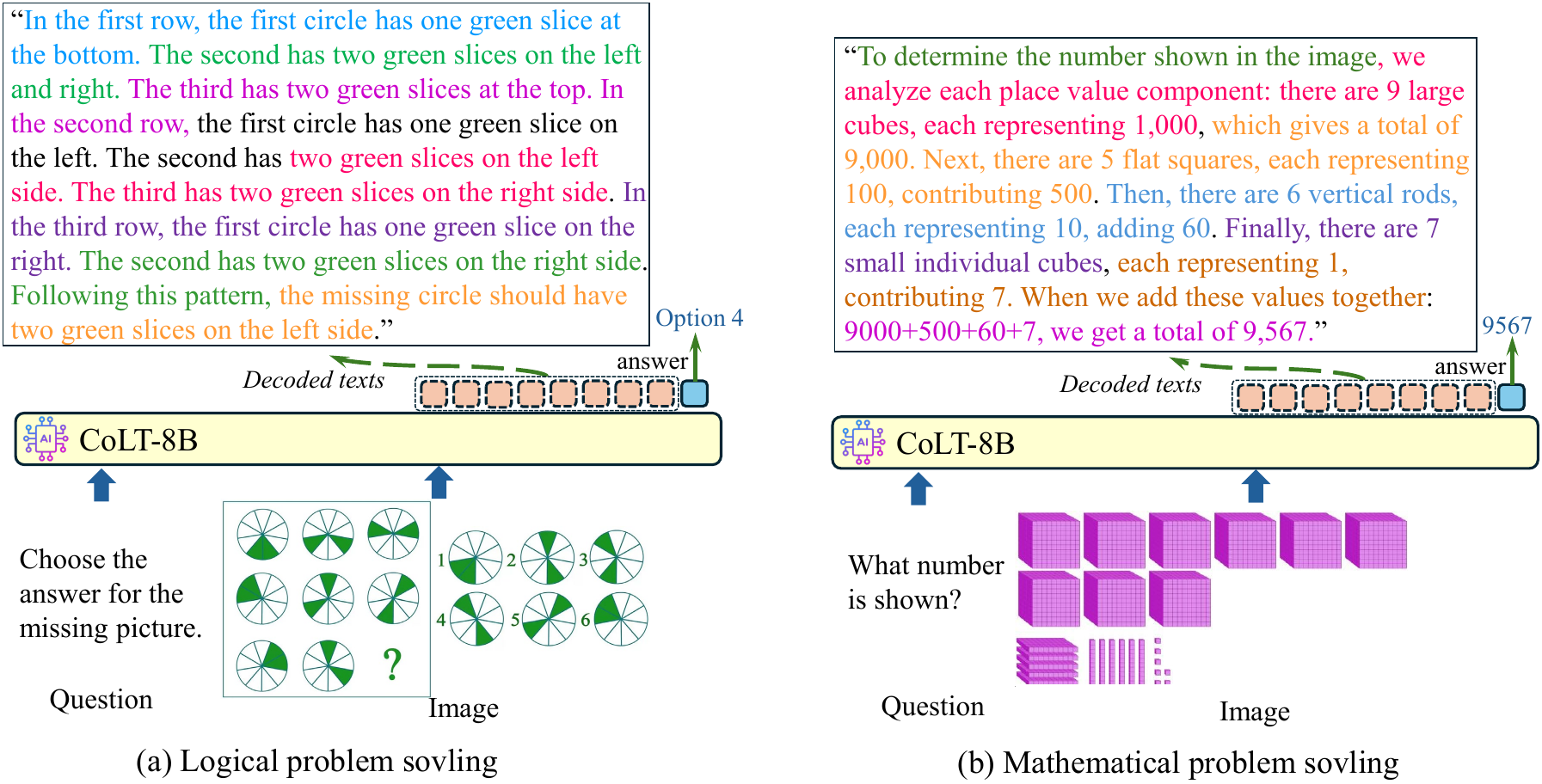}}
    %\fbox{\parbox{0.95\linewidth}{\centering\vspace{3cm}\textbf{[Qualitative Comparison]}\\\vspace{0.3cm}\small Two example problems (one mathematical, one scientific) showing: (a) the input image and question, (b) text CoT reasoning with explicit steps, (c) decoded latent thoughts from \ours{} showing more concise reasoning patterns, and (d) the final answers.\vspace{0.5cm}}}

    \caption{Two qualitative examples from MathVista with decoded latent thoughts. We use the forward decoder to project latent states back into text. Color-coded segments indicate reasoning content decoded from different latent steps. }
%(a)~A logical pattern-matching problem where latent steps capture row-by-row visual analysis and pattern inference. (b)~A mathematical problem where latent steps decompose place-value components and compute the sum. Both examples confirm that the latent states encode meaningful, structured reasoning.
    \label{fig:qualitative}

\end{figure}

\subsection{Qualitative Analysis}
\Cref{fig:qualitative} presents two MathVista examples where we use the forward decoder to project the latent thoughts of \ours{} back into text for interpretation. In the logical pattern-matching problem~(a), the decoded text reveals that the latent steps systematically describe the visual pattern row by row, identifying the position and number of green slices in each circle, and then infer the missing pattern to arrive at the correct answer. In the mathematical problem~(b), the decoded text shows that the latent steps decompose the base-10 block representation into individual place-value components and compute the final sum. The color-coded segments demonstrate that each latent step encodes a distinct and meaningful portion of the reasoning chain, confirming that \ours{}'s latent representations capture structured, interpretable thought patterns. The decoded content also maintains logical coherence across consecutive latent steps, meaning that our internal supervision enforces meaningful step-to-step transitions.

%\Cref{fig:qualitative} presents two MathVista examples where we use the forward decoder to project the latent thoughts of \ours{} back into text for interpretation. In the logical pattern-matching problem~(a), the decoded text reveals that the latent steps systematically describe the visual pattern row by row, identifying the position and number of green slices in each circle, and then infer the missing pattern to arrive at the correct answer. In the mathematical problem~(b), the decoded text shows that the latent steps decompose the base-10 block representation into individual place-value components and compute the final sum. The color-coded segments demonstrate that each latent step encodes a distinct and meaningful portion of the reasoning chain, confirming that \ours{}'s latent representations capture structured, interpretable thought patterns. These visualizations further validate that the three-way supervision effectively guides latent thoughts toward encoding structured semantics rather than degenerating into arbitrary representations. The decoded content also maintains logical coherence across consecutive latent steps, suggesting that the internal supervision successfully enforces meaningful step-to-step transitions.

%=============================================
\section{Conclusion}
\label{sec:conclusion}

We presented \ours{} (Chain of Latent Thoughts), a novel framework that teaches multi-modal models to reason through latent thought representations instead of explicit text-based CoT. The core innovation is introducing three complementary step-level supervisions: forward decoding, backward decoding, and internal supervision. Extensive experiments on eight benchmarks demonstrate that \ours{} consistently outperforms existing latent reasoning methods and achieves up to $22.6{\times}$ speedup over text CoT. Promising directions include adaptive latent steps by problem difficulty, and exploring hybrid latent-text reasoning frameworks. % extending latent reasoning to video and audio modalities,

\section*{Acknowledgements}
Our work is supported by National Natural Science Foundation of China (Grant No. U2574216), and Emerging Frontiers Cultivation Program of Tianjin University Interdisciplinary Center. This research is also supported by the Fundamental Research Funds for the Central Universities (No. XXX-63263254); by the National Research Foundation Singapore and the Cyber Security Agency under the National Cybersecurity R\&D Programme (NCRP25-P04-TAICeN); and is part of the IN-CYPHER programme and is supported by the National Research Foundation, Prime Minister's Office, Singapore under its Campus for Research Excellence and Technological Enterprise (CREATE) programme. Any opinions, findings and conclusions, or recommendations expressed in these materials are those of the author(s) and do not reflect the views of the National Research Foundation, Singapore, Cyber Security Agency of Singapore, Singapore.
\clearpage
% ---- Bibliography ----
\bibliographystyle{splncs04}
\bibliography{main}
\end{document}